\documentclass[final, conference, a4paper]{IEEEtran}

\usepackage{amsmath}   
\usepackage{amssymb}
\usepackage{amsfonts}  
\usepackage{graphicx}  
\usepackage{url}       
\usepackage[noadjust]{cite}
\usepackage{citesort}   
\usepackage{prettyref}
\usepackage{balance}
\usepackage{subfig}   
\usepackage{xr}
\usepackage{multirow}
\usepackage{standalone}
\usepackage{pgf}
\usepackage{pgfplots}
\usepackage{verbatim}
\usepackage{tikz}
\usepackage{array}
\usepackage{booktabs}
\usepackage{xcolor}
\usetikzlibrary{arrows.meta,positioning, shapes, patterns, calc, through}
\tikzstyle{vertex}=[circle, draw, inner sep=1pt, minimum size=6pt]
\pgfplotsset{compat=1.17}
\usepackage[
single=true,
sort=true,
only-used=true
]{acro}
\DeclareAcronym{BLE}{
	short = BLE,
	long = Bluetooth-Low-Energy
}

\DeclareAcronym{UWB}{
	short = UWB,
	long = Ultra-Wideband
}

\DeclareAcronym{TWR}{
	short = TWR,
	long = Two-Way-Ranging
}

\DeclareAcronym{EKF}{
	short = EKF,
	long = Extended Kalman Filter
}

\DeclareAcronym{IoT}{
	short = IoT,
	long = internet of things
}

\DeclareAcronym{LOS}{
	short = LoS,
	long = Line-of-Sight
}

\DeclareAcronym{NLOS}{
	short = NLoS,
	long = Non-Line-of-Sight
}

\DeclareAcronym{RTOS}{
	short = RTOS,
	long = Real-Time Operating System
}

\DeclareAcronym{TOF}{
	short = TOF,
	long = time-of-flight
}

\DeclareAcronym{TdoA}{
	short = TdoA,
	long = Time-difference-of-Arrival
}

\DeclareAcronym{PCB}{
	short = PCB,
	long = printed circuit board
}

\DeclareAcronym{EEPROM}{
	short = EEPROM,
	long = electric erasable programmable read-only memory
}

\IEEEoverridecommandlockouts 

\begin{document}

\renewcommand{\baselinestretch}{0.97} 

\title{Ultra-Wideband (UWB) Positioning System Based on ESP32 and DWM3000 Modules}

\author{
	\authorblockN{Sebastian Krebs}
	\authorblockN{University of Applied Sciences HTWG Konstanz}
  	\authorblockN{Alfred-Wachtel-Str.~8, 78462 Konstanz, Germany}
  	\authorblockA{Email: se391kre@htwg-konstanz.de}
	\authorblockA{sebi.krebs@web.de}
	\and
	\authorblockN{Tom Herter}
	\authorblockN{University of Applied Sciences HTWG Konstanz}
  	\authorblockN{Alfred-Wachtel-Str.~8, 78462 Konstanz, Germany}
  	\authorblockA{Email: to351her@htwg-konstanz.de}
	\authorblockA{tom28.06@gmx.de}
}

\maketitle

\begin{abstract}
  In this paper, an \ac{UWB} positioning system is introduced, that leverages six identical custom-designed boards, each featuring an ESP32 microcontroller and a DWM3000 module from Quorvo.
  \newline
  The system is capable of achieving localization with an accuracy of up to 10\,cm, by utilizing \ac{TWR} measurements between one designated ''tag'' and five ''anchor'' devices.
  The gathered distance measurements are subsequently processed by an \ac{EKF} running locally on the tag board, enabling it to determine its own position,
  relying on fixed, a priori known positions of the anchor boards.
  \newline
  This paper presents a comprehensive overview of the system's architecture, the key components, and the capabilities it offers for indoor positioning and tracking applications.
\end{abstract}


\IEEEpeerreviewmaketitle

\section{Introduction}
\label{sec:Introduction}

Indoor positioning and tracking systems gain importance in a variety of industrial fields as well as in research \cite{LiFi_Positioning,Accelerometer_Positioning,UWB_Positioning,UWB_Positioning2}.

Traditional positioning systems, however, often encounter certain limitations, such as accuracy to the meter \cite{LiFi_Positioning} or an increased positioning deviation \cite{Accelerometer_Positioning}. 

To achive an improved accuracy and a reduced variance Ultra-Wideband (UWB) positioning systems have been applied to various scenarios \cite{UWB_Positioning} \cite{UWB_Positioning2}.
Building on this technology, we have developed an \ac{UWB} positioning system that utilizes hardware and advanced algorithms to generate precise position information in the three-dimensional space.
However, the estimations that are conducted over the course of its development are limited to the two-dimensional space. 

The designed \ac{UWB} system consists of six identical boards that are all based on the ESP32 microcontroller, a versatile and powerful platform that is often used in low-cost \ac{IoT} scenarios \cite{ESP32}. 
These custom-made \acp{PCB} \footnote{Layout and production files are open-source available \cite{uwb-tracking}.} are equipped with DWM3000 modules from Quorvo, which utilizes \ac{UWB} functionalities for short-range wireless communication.

One of these boards is referred to as a ''tag''.
It is responsible for initiating measurements with the other five "anchor" boards.
The innovative aspect of the system lies in its ability to perform accurate localization without dependence on external infrastructure for its processing,
since all the necessary calculations are performed on the device itself that is being localized.

The heart of the positioning system is an \acf{EKF} implemented locally on the tag board.
This \ac{EKF} takes the distance measurements obtained by \acf{TWR} with the anchor boards
and based on their a priori known positions, calculates the real-time position of the tag board.

The system is partially scalable; the number of anchors can be significantly increased at
the expense of the general round-trip time.
The ratio is linear, an increaing anchor amount results in a longer roundtrip time.

The measuring principle of distance measurement is explained in the following section.
A distance measurement was implemented based on the associated IEEE standard  \cite{IEEE802154a} \cite{IEEE802154z}.
The chapter \ref{Section:principle} explains the system architecture,
including the choice of anchor positions and 
information about the scheduled timing in relation to position measurements.
This is followed by a brief description of the designed hardware and
its wide range of applications as a general evaluation board.
In order to explain the implementation of the firmware in more detail,
chapter \ref{section:firmware} describes the distribution of functionalities to various tasks of the \ac{RTOS}.
The results of static tests can be found in chapter \ref{section:tests}.
Both, the spatial resolution and the most important limitations are explained.
Finally, the last chapter summarizes the findings and provides a critical review of the results.

\section{Measurement Priciple}\label{Section:principle}
\acf{TWR} is a foundational technique for obtaining precise distance measurements within the \ac{UWB} positioning system.
It relies on the time it takes for signals to propagate from a tag board to an anchor board and back again.
This time measurement, in compliance with the IEEE 802.15.4a/4z standards\cite{IEEE802154a}\cite{IEEE802154z}, offers the basis for distance estimation by multiplying the time traveled with the speed of light.

The following figure \ref{fig:twr} shows how a \ac{TWR} handshake takes place.
The tag's firmware calculates the \acf{TOF} as well as the distance between both devices by comparing the timestamps of sending and reception. 
Therefore it is necessary for the \ac{UWB} messages to not only be tracked by the time they are received at the tag, but also to contain information about when each anchor received it and when it starts to transmit the response. 

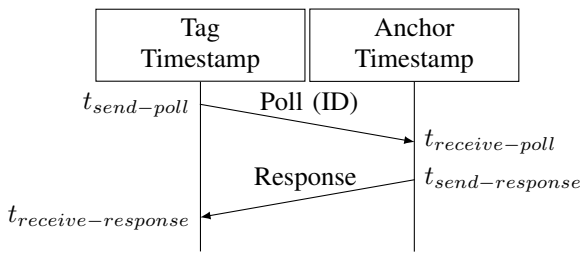
\begin{figure}[hbt!]
	\centering
	\begin{tikzpicture}[
		node distance=1.5cm,
		block/.style={
			align=center, draw,minimum width=2.75cm, minimum height=.5cm
		},
		ghost/.style={
			align=center,minimum width=0cm, minimum height=0cm
		},
		label/.style={
			minimum width=2cm, minimum height=0cm, text width=2.5cm
		},
		>=latex,
		]
		\node [block] (tag_timestamp) {Tag\\Timestamp};
		\node [block, right=.05cm of tag_timestamp] (anchor_timestamp) {Anchor\\Timestamp};
		
		\node [ghost, below=2.25cm of tag_timestamp] (tag_endpoint) {};
		\node [ghost, below=2.25cm of anchor_timestamp] (anchor_endpoint) {};
		
		\draw [-] (tag_timestamp) -- (tag_endpoint);
		\draw [-] (anchor_timestamp) -- (anchor_endpoint);
		
		\node [label, align=right, anchor=east, below=.5cm of tag_timestamp.west] (t_send_poll) {$t_{send-poll}$};
		\node [label, align=right, anchor=east, below=2cm of tag_timestamp.west] (t_receive_response) {$t_{receive-response}$};
		
		\node [label, align=left, anchor=west, below=1cm of anchor_timestamp.east] (t_receive_poll) {$t_{receive-poll}$};
		\node [label, align=left, anchor=west, below=1.5cm of anchor_timestamp.east] (t_send_response) {$t_{send-response}$};
		
		\draw [->] (t_send_poll.east) -- (t_receive_poll.west);
		\draw [->] (t_send_response.west) -- (t_receive_response.east);
		\node [ghost, below=.5cm of anchor_timestamp.west] (poll_label) {Poll (ID)};
		\node [ghost, below=1.5cm of tag_timestamp.east] (response_label) {Response};
		
	\end{tikzpicture}
	\caption{Timing diagram of \acf{TWR}}
	\label{fig:twr}
\end{figure}

\begin{figure}[hbt!]
	\centering
	\begin{tikzpicture}
	\begin{axis}[
		ybar=0pt,
		bar width=10pt,
		bar shift=5pt,
		xlabel={Time [ms]},
		ylabel={Anchor index},
		ymin=0, ymax=6,
		xmin=0, xmax=500,
		legend style={at={(0.83,0.97)},
		anchor=north,legend columns=1},
		ymajorgrids=true,
		grid style=dashed,
		]
		\addplot table[col sep=comma, x=Data, y=Poll]{timing.csv};
		\addplot table[col sep=comma, x=Data, y=Response]{timing.csv};
		\legend{Poll, Response}

        \draw [thick=4pt] (axis cs:250,\pgfkeysvalueof{/pgfplots/ymin}) -- (axis cs:250,\pgfkeysvalueof{/pgfplots/ymax});
        \node at (axis cs:250,\pgfkeysvalueof{/pgfplots/ymin}+3) [below, rotate=90] {round-trip time};
    
	\end{axis}
	\end{tikzpicture}
	\caption{Timing diagram of the \ac{TWR} measurements.}
	\label{fig:csvbarchart}
\end{figure}
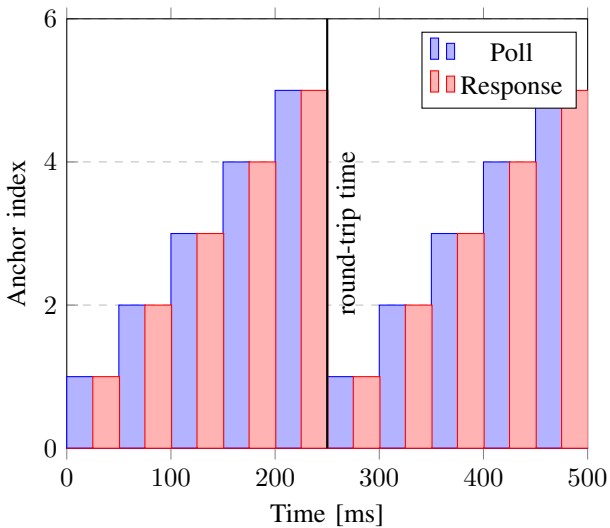

In figure \ref{fig:csvbarchart} illustrates the timing of one positioning cycle.
During the static tests conducted in section \ref{section:tests} five anchor devices were utilized.
The total time for position estimation is set to be 250 ms, allocating 50 ms for each distance measurement.
The 250 ms round-trip time is a result of anchor amount chosen.
The duration for each distance measurement, or ranging time, is set at a firm 50 ms due to limitations imposed by the processing logic.
As a result, the total round-trip time is affected by the number of anchor devices utilized.
Adding more anchors increases the amount of individual \acp{TWR}, leading to a longer round-trip time.

For detailed technical specifications and methods, interested readers are referred to the documentation of the IEEE 802.15.4a/4z standards \cite{IEEE802154a} \cite{IEEE802154z}, which provides comprehensive guidelines for the orchestration of \ac{UWB} signals and calculation of \ac{TOF}.
These standards ensure the correctness and accuracy of our distance measurements.

\section{System Architecture}\label{section:system_arch}
In the presented scenario, five anchors are strategically distributed throughout the room,
positioned at a height of 3\,m just under the ceiling to maximize the likelihood of 
\ac{LOS} conditions.
This is due to the fact that it facilitates more accurate \ac{TOF} measurements, by minimizing interference caused by multipath propagation, and enhance signal reliability,
leading to more precise and reliable distance calculation than in an \ac{NLOS} environment.
These five anchors do not rely on any information specific to their mounting position.

During the setup of the system, anchor positions are being transmitted to the tag via a \ac{BLE} interface.
The tag's Firmware stores these configurations in persistent \ac{EEPROM} to ensure their availability throughout power cycles.
The tag leverages these provided anchor positions, in conjunction with distance measurements, to determine its own position. 


\begin{figure}[hbt!]
	\centering
	\includegraphics[scale=0.08]{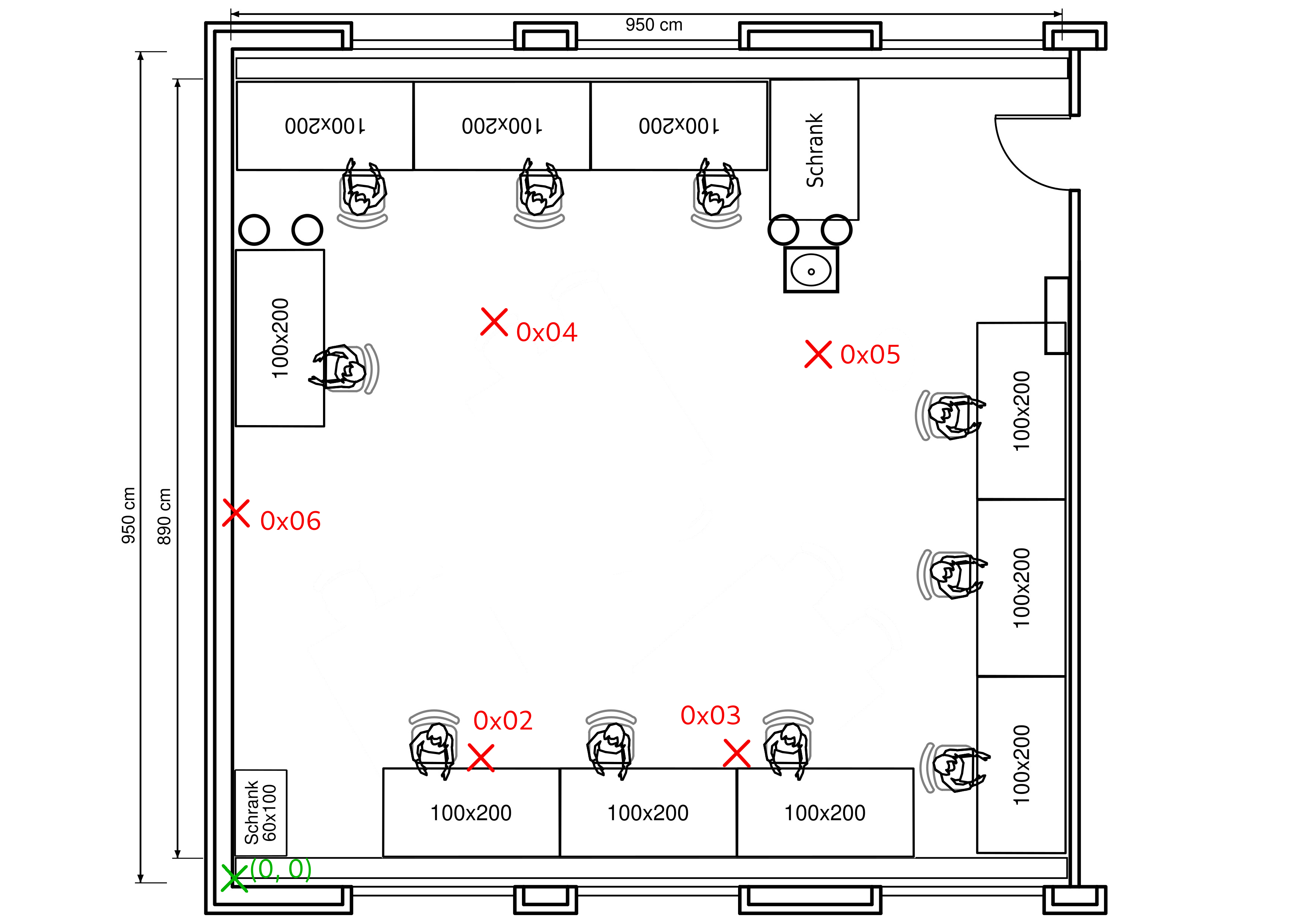}
	\caption{System architecture for positioning.}
	\label{fig:systemarch}
\end{figure}

\begin{table}[hbt!]
	\centering
	\begin{tabular}{l l l l c}
		\textbf{anchor} & \textbf{x-coordinate[m]} & \textbf{y-coordinate[m]} & \textbf{z-coordinate[m]}\\
		0x02 & $0.81$m & $3.63$m & $3.01$m\\
		0x03 & $0.81$m & $6.38$m & $3.01$m\\
		0x04 & $6.31$m & $7.66$m & $2.83$m\\
		0x05 & $6.72$m & $3.65$m & $2.64$m\\
		0x06 & $2.77$m & $0.07$m & $0.91$m\\
		
	\end{tabular}
	\caption{Coordinates of the anchors.}
	\label{table:anchor_positions}
\end{table}

As can be seen from the illustration of the room, care was taken to distribute the anchors evenly.
Figure \ref{fig:systemarch} also shows that the distribution of anchors is also distributed in the z axis.
However, more attention was paid to an overhanging installation in order to reduce \ac{NLOS} conditions.

The time for one \ac{TWR} could be reduced to 50\,ms per bidirectional recurrence, without sacrificing accuracy,
leading to a general position update interval of 250\,ms with five anchors. 
This low ranging time of 50\,ms was achieved by implementing the firmware in a interrupt-based manner for the \ac{UWB} message recognition, and utilizing the ESPs multi-core processing capabilities.

Detailed implementation-specific information about the chosen procedures to optimize the round-trip time by reducing each individual ranging time
is given in section \ref{section:firmware}.

\section{Hardware Design}\label{section:hardware}
The hardware design of \ac{UWB} boards draws inspiration from pre-existing DWM3000 evaluation
boards.
However, the proprietary board development enables specialized component selection, tailored to their intended use cases.
User-friendliness was a paramount consideration during the design process,
resulting in the integration of multiple user buttons and indicating LEDs for versatile purposes like triggering the initialization of the \ac{BLE} server, in order to view or change the stored anchor positions. 
Additionally, the PCB incorporates a convenient on-board LiPo battery charging and protection circuit via USB-C interface.

\begin{figure}[hbt!]
	\centering
	\includegraphics[scale=0.30]{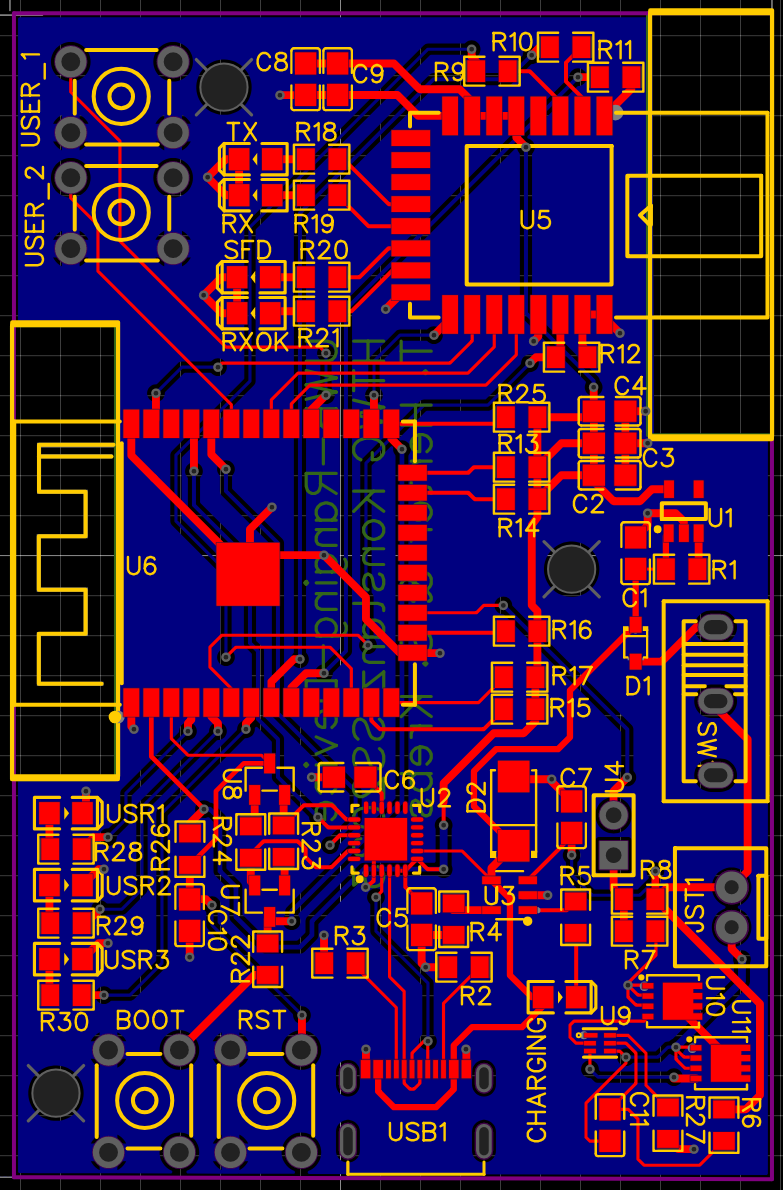}
	\caption{PCB layout of the UWB board.}
	\label{fig:pcb}
\end{figure}
As shown in figure \ref{fig:pcb}, the board is designed in a compact form,
making it suitable for evaluation and development purposes.
Notably, our PCB design ensures a consistent layout across all boards, regardless of their specific application.
Below the antennas of both the DWM3000 and ESP32, the ground filling has been selectively omitted, to ensure antenna radiation characteristics according to the corresponding data sheets, enabling more precise measurements.

\section{Firmware Architecture}\label{section:firmware}
To meet the demanding timing requirements essential for \ac{TOF} measurements and enhance network round-trip times,
the firmware is based on a \ac{RTOS}.
More precisely FreeRTOS \cite{FreeRTOS_2023} offers the capability to create multiple tasks, that collaborate coherently and can be executed simultaneously on each of the two cores of the ESP32 microcontroller.

Documentation of the source code is continuously made available online with the help of Doxygen and GitHub pages \cite{doxygen-doku}.
The implementation can be viewed publicly on GitHub \cite{uwb-tracking}.

Both the firmware of the tag and anchor execute the \ac{TOF} task.
This utilizes an inheritance-based implementation that sets the distinction between \ac{TWR} initiator or \ac{TWR} responder of the ranging measurement.
When a device is configured as a tag,
it additionally undertakes the execution of \ac{EKF} task.
The \ac{EKF} task processes the distance measurements generated by the \ac{TOF} task,
culminating in a positional estimation.

\subsection{TOF-Task}\label{section:firmware-tof}
The \ac{TOF} task is based on the example code
provided by Quorvo for \ac{TWR} measurements.
However, we have refined the functionality by structuring in a class-based framework.
The commonalities between initiator and responder roles have been
encapsulated in a common superclass.
This design decision allows us to maintain a lean and clear software structure,
reduce redundancy and simplify maintenance.

In practice, the tag, which acts as the \ac{TWR} initiator, as part of the localization architecture iterates through a list of anchors, each of which serves as an individual \ac{TWR} responder. 
The tag, jointly with the anchor, generates distance measurements. 
The result is a comprehensive data set that serves as the basis for coordinate calculation. 

\subsection{EKF-Task}\label{section:firmware-ekf}
The \ac{EKF} employed in our system leverages two distinct mathematical models
to achieve position estimation.
For readers interested in delving deeper into the theoretical foundations of the
Kalman Filter, we recommend consulting the work of Li Qiang et al. in
"Kalman Filter and Its Application"~\cite{Kalman}.
The \ac{EKF} utilizes a prediction model based on the constant velocity and trajectory model assumption.

This model serves as a fundamental tool for estimating the next position of the tag.
For potentially more dynamic systems the use of nonlinear models
for tag movement could improve the dynamic behavior of the Kalman Filter.

The measurement model is encapsulated in the Jacobian matrix given in equation~\ref{eq:measurementmatrix},
enabling the transformation of measured distances into accurate position estimations. 
The measurement model finds its expression in measurement matrix $H$, 
which effectively links the measured distances to position estimation:

\begin{equation}
	\begin{aligned}
		\Delta x_i &= x_i - x \\
		\Delta y_i &= y_i - y \\
		\Delta z_i &= z_i - z \\
		dist_i &= \sqrt{{\Delta x_i^2 + \Delta y_i^2 + \Delta z_i^2}} \\
		H &= \begin{bmatrix}
			-\frac{{\Delta x_1}}{{dist_1}} & -\frac{{\Delta y_1}}{{dist_1}} & -\frac{{\Delta z_1}}{{dist_1}} \\
			-\frac{{\Delta x_2}}{{dist_2}} & -\frac{{\Delta y_2}}{{dist_2}} & -\frac{{\Delta z_2}}{{dist_2}} \\
			\vdots & \vdots & \vdots \\
			-\frac{{\Delta x_{\text{max}}}}{{dist_{\text{max}}}} & -\frac{{\Delta y_{\text{max}}}}{{dist_{\text{max}}}} & -\frac{{\Delta z_{\text{max}}}}{{dist_{\text{max}}}}
		\end{bmatrix}
	\end{aligned}
	\label{eq:measurementmatrix}
\end{equation}

By using matrix $H$, the \ac{EKF} is able to directly translate position estimate into predicted distance measurements.
These are further compared to real distance measurements and based on deviation between these, position estimate is improved.

\section{Test Results}\label{section:tests}
The implementation of localization systems often requires evaluation of measured values through empirical,
static and dynamic tests.
Since the dynamics of position determination are largely determined by parameterization of \ac{EKF},
a detailed evaluation of the system's ability to detect moving objects is not examined in this paper.

The effect of anchor arrangement is crucial to the success of the system.
In case of these tests, most anchors were mounted at a height of 4m.
One single anchor was assembled at a height of 1m to ensure good position resolution in the z axis.
Otherwise, an overestimation of distances and, thus, a poor calculation of the z axis was determined.

During testing, attention was paid to keep \ac{LOS} conditions with all anchors,
although tests showed that the influence of \ac{NLOS} conditions can be partially compensated by \ac{EKF}.
The table \ref{table:measurements} shows fixed positions in the room that were measured using the system
over a fixed period of time in a grid-like pattern. 
To do this, mean deviation and fluctuation of the values are evaluated. 
Before each test sequence it is ensured, that the \ac{EKF} is not biased by values of previous measurements. 

\begin{table}[hbt!]
	\centering
	\begin{tabular}{l l l c}
		\textbf{Position[m]} & \textbf{$\mu$[cm]} & \textbf{$\sigma$[cm]}\\
		$(x,y)$ & $mean(errors)$ & $std(errors)$\\
		& ''accuracy'' & ''precision''\\
		$(2,2)$ & $67.66$ cm & $2.90$ cm\\
		$(2,3)$ & $55.42$ cm & $2.97$ cm\\
		$(2,4)$ & $65.05$ cm & $2.8$ cm\\
		$(2,5)$ & $47.31$ cm & $3.51$ cm\\
		$(2,6)$ & $46.07$ cm & $3.13$ cm\\
		$(2,7)$ & $54.00$ cm & $3.17$ cm\\
		$(2,8)$ & $34.48$ cm & $3.3$ cm\\

		$(3,2)$ & $31.74$ cm & $3.42$ cm\\
		$(3,3)$ & $121.37$ cm & $4.47$ cm\\
		$(3,4)$ & $23.06$ cm & $3.07$ cm\\
		$(3,5)$ & $29.34$ cm & $2.91$ cm\\
		$(3,6)$ & $28.31$ cm & $2.93$ cm\\
		$(3,7)$ & $22.68$ cm & $3.2$ cm\\
		$(3,8)$ & $32.98$ cm & $3.73$ cm\\

		$(4,2)$ & $19.75$ cm & $2.97$ cm\\
		$(4,3)$ & $21.73$ cm & $3.4$ cm\\
		$(4,4)$ & $29.91$ cm & $2.9$ cm\\
		$(4,5)$ & $9.27$ cm & $3.2$ cm\\
		$(4,6)$ & $28.1$ cm & $2.84$ cm\\
		$(4,7)$ & $35.95$ cm & $8.34$ cm\\
		$(4,8)$ & $14.88$ cm & $3.13$ cm\\

		$(5,2)$ & $80.12$ cm & $3.61$ cm\\
		$(5,3)$ & $61.88$ cm & $4.19$ cm\\
		$(5,4)$ & $47.44$ cm & $3.57$ cm\\
		$(5,5)$ & $21.28$ cm & $6.76$ cm\\
		$(5,6)$ & $31.89$ cm & $3.65$ cm\\
		$(5,7)$ & $33.02$ cm & $3.72$ cm\\
		$(5,8)$ & $26.27$ cm & $3.73$ cm\\

		$(6,2)$ & $89.12$ cm & $2.99$ cm\\
		$(6,3)$ & $99.28$ cm & $4.28$ cm\\
		$(6,4)$ & $73.01$ cm & $3.88$ cm\\
		$(6,5)$ & $55.42$ cm & $6.16$ cm\\
		$(6,6)$ & $10.63$ cm & $4.56$ cm\\
		$(6,7)$ & $29.46$ cm & $7.01$ cm\\
		$(6,8)$ & $13.0$ cm & $2.86$ cm\\
		
	\end{tabular}
	\caption{Static measurement deviation out of 500 measurements at each grid intersection point.}
	\label{table:measurements}
\end{table}

\begin{figure}[hbt!]
	\includegraphics[scale=0.27]{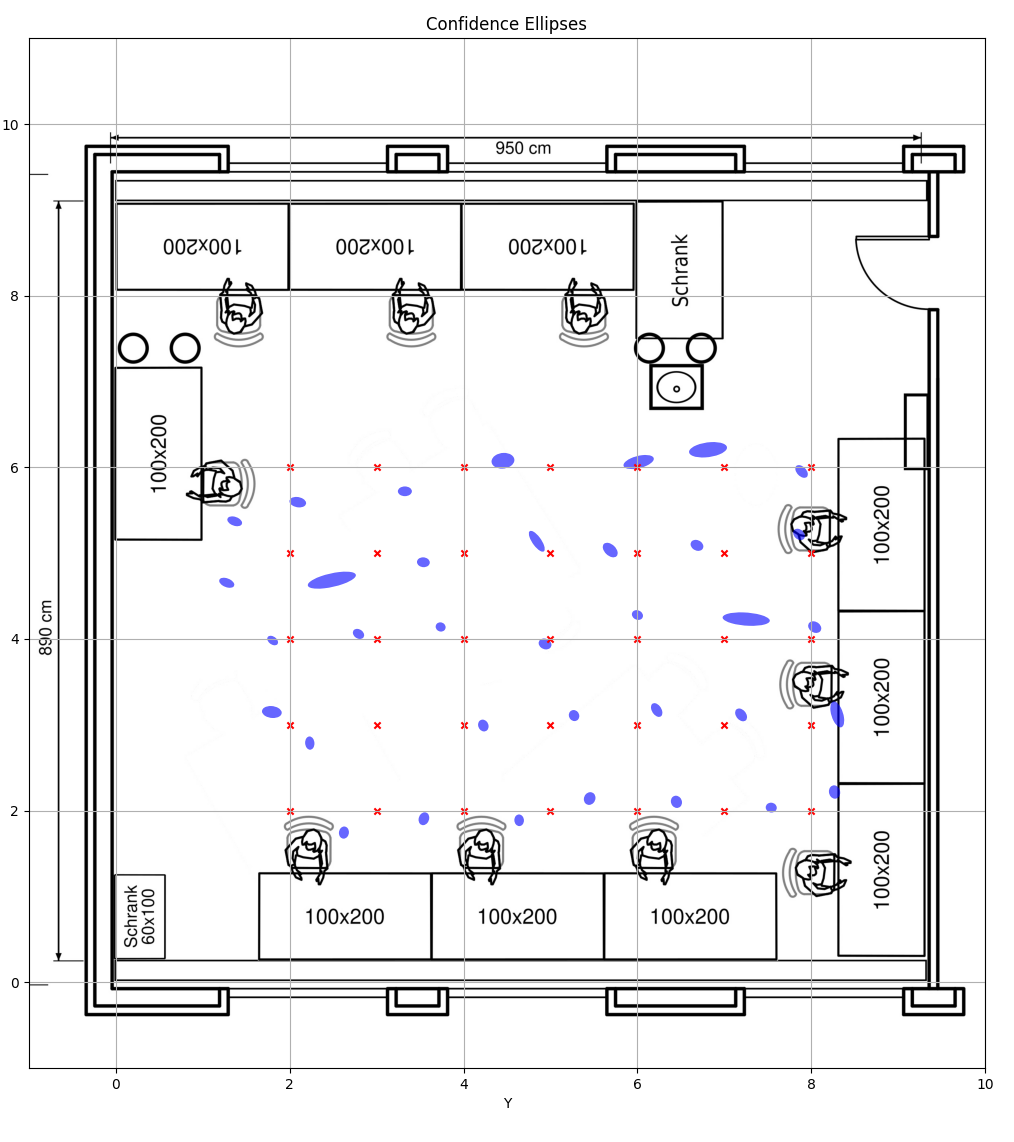}
	\caption{Confidence ellipses of the grid measurement with 99.7\,\% of the points inside of the ellipse are shown in blue.
	Displayed in red is the corresponding real position.}
	\label{fig:statistics}
\end{figure}

The 10\,m\,x\,8\,m room was divided into 1\,m\,x\,1\,m squares. 
Multiple measurements were then carried out over the course of three minutes at each grid intersection point.
This procedure allows deviations of positions to be represented per location in the room.
For illustration in figure \ref{fig:statistics}, $3\sigma$-ellipses were drawn to show the scattering per position.

The graph in figure \ref{fig:statistics} shows, that the biggest part of the scattering occurs in x-direction. 
A systematic error seems to appear when taking a look at the deviation in y-direction. 
The more distance is between the measured intersection point and the center of the room (4\,m\,x\,5\,m), the larger the deviation of the mean is to actual positions. 
Furthermore, it seams that the anchor placement was not well distributed for measurements in the area for high x and low y coordinates. 
A big deviation can be observed at the coordinates 3\,m\,x\,3\,m which could be validated through multible measurements.
A possible reason for that could be good conditions for multipath propagation of \ac{UWB} signals.
Overall the best performance is achieved in direct center of the grid where distance to every anchor is roughly equal.


\section{Conclusion And Outlook}\label{section:conclusion}
In conclusion, it can be said that a relatively precise localization system has been designed.
Both the hardware and the software were designed specifically for use as a positioning module
and fulfill this purpose with a small variance in positioning.

As can be seen from the grid-based measurement, the system achieves a variance of just 3 - 6cm.
As can be seen also in this measurement,
the offset of position determination is strongly dependent on position in space
and relative position of anchors.
Studies such as those by Zhao et. al.~\cite{Zhao_2022} show that the anchor positions
have a significant influence on the positioning accuracy.
This effect could be confirmed again with this implementation.

A critical point is the system's scalability, the number of anchors has a linear impact on its temporal performance.
By pinging every single anchor, the tag is not able to handle a very large number of anchors without increasing the roundtrip time.
This problem could be avoided if instead of \ac{TWR} measurements, \ac{TdoA} measurements would be performed.
Even if this measurement principle requires a nanosecond precise synchronization of the anchors,
roundtrip time would be limited to the duration of one ping process.

Initially, a hybrid solution using \ac{TWR} and \ac{TdoA} measurements was planned,
but the wireless clock synchronization of the anchors is currently only possible with limited accuracy.
Further research in this area could allow the system to generate position measurements with a roundtrip time of up to 50 ms.


\bibliographystyle{IEEEtran} \bibliography{boid}

\end{document}